\pdfoutput=1

\documentclass[10pt,logo,copyright]{arxiv/nvidiatechreport}
\usepackage[authoryear,sort&compress,round]{natbib}

\usepackage[utf8]{inputenc} %
\usepackage[T1]{fontenc}    %

\usepackage{parskip}        %
\usepackage{url}            %
\usepackage{hyperref}       %
\usepackage{booktabs}       %
\usepackage{nicefrac}       %
\usepackage{microtype}      %
\usepackage{graphicx}
\usepackage{animate}        %
\usepackage{subcaption}
\usepackage{tabularx}
\usepackage{makecell}
\usepackage{adjustbox}
\usepackage{setspace}
\newcolumntype{M}[1]{>{\centering\arraybackslash}m{#1}}
\usepackage{float}
\usepackage{afterpage}
\usepackage{tikz}
\usetikzlibrary{positioning,shapes,arrows}
\usepackage{amsmath,amsfonts,bm, bbm,leftindex}
\usepackage{multirow}
\usepackage{comment}
\usepackage{gensymb}
\usepackage{lipsum}
\usetikzlibrary{arrows.meta, positioning, fit}
\usepackage[para]{threeparttable}
\usetikzlibrary{tikzmark}

\let\cite\citep

\graphicspath{{./}{figures/}}

\definecolor{darkred}{rgb}{0.7, 0.0, 0.0}

\usepackage{pifont}

\usepackage{wrapfig}

\usepackage[nameinlink]{cleveref}
\crefname{equation}{Eq.}{Eqs.}
\crefname{figure}{Fig.}{Figs.}
\crefname{section}{Sec.}{Sec.}
\crefname{appendix}{App.}{App.}
\crefname{table}{Tab.}{Tabs.}
\crefname{algorithm}{Algo}{Algo}
\crefname{thm}{Thm}{Thm}
\Crefname{thm}{Thm}{Thm}
\crefname{prop}{Prop}{Prop}
\usepackage{longtable}

\usepackage{colortbl}
\usepackage{array}

\usepackage{mathabx}

\usepackage{xspace}
\makeatletter
\DeclareRobustCommand\onedot{\futurelet\@let@token\@onedot}
\def\@onedot{\ifx\@let@token.\else.\null\fi\xspace}
\def\eg{\emph{e.g}\onedot}

\makeatother

\usepackage{amsfonts}
\usepackage{bm}
\usepackage{amsmath}
\usepackage{mathtools}
\usepackage{multirow}
\usepackage{booktabs}
\usepackage{caption}
\usepackage{enumitem}
\usepackage{wrapfig,lipsum,booktabs}
\usepackage{caption}
\usepackage{bbm}
\usepackage{multirow}
\usepackage{pifont}
\usepackage[linesnumbered,ruled,vlined]{algorithm2e}

\SetCommentSty{mycommfont}
\SetAlCapFnt{\small}
\usepackage{etoc}
\SetAlgorithmName{Algo}{Algo}{}

\renewcommand{\paragraph}[1]{{\vspace{1mm}\noindent \bf #1}.}

\newcommand{\METHOD}{GRAIL\xspace}
\newcommand{\link}{https://research.nvidia.com/labs/dair/grail/}

\title{\METHOD: Generating Humanoid Loco-Manipulation from World Models}

\author{
    Tianyi~Xie$^{1,2,\dagger}$, Haotian~Zhang$^{1,\dagger}$, Jinhyung~Park$^{1,\dagger}$, Zi~Wang$^{1,\dagger}$, Bowen~Wen$^{1}$,
    Jiefeng~Li$^{1}$, Xueting~Li$^{1}$, Qingwei~Ben$^{1}$, Haoyang~Weng$^{1}$, Yufei~Ye$^{1}$, David~Minor$^{1}$,
    Tingwu~Wang$^{1}$, Chenfanfu~Jiang$^{2}$, Sanja~Fidler$^{1}$, Jan~Kautz$^{1}$, Linxi~Fan$^{1}$, Yuke~Zhu$^{1}$,
    Zhengyi~Luo$^{1,\ddagger}$, Umar~Iqbal$^{1,\ddagger}$, Ye~Yuan$^{1,\ddagger}$ \\
    \small$^{1}$ NVIDIA \quad $^{2}$ UCLA \\
    \small$^\dagger$ Co-First Authors \quad $^\ddagger$ Project Leads \\
    \small \tt{\href{\link}{\link}}
}

\begin{document}
\maketitle

\begin{figure}[th]
    \centering
    \includegraphics[width=\textwidth]{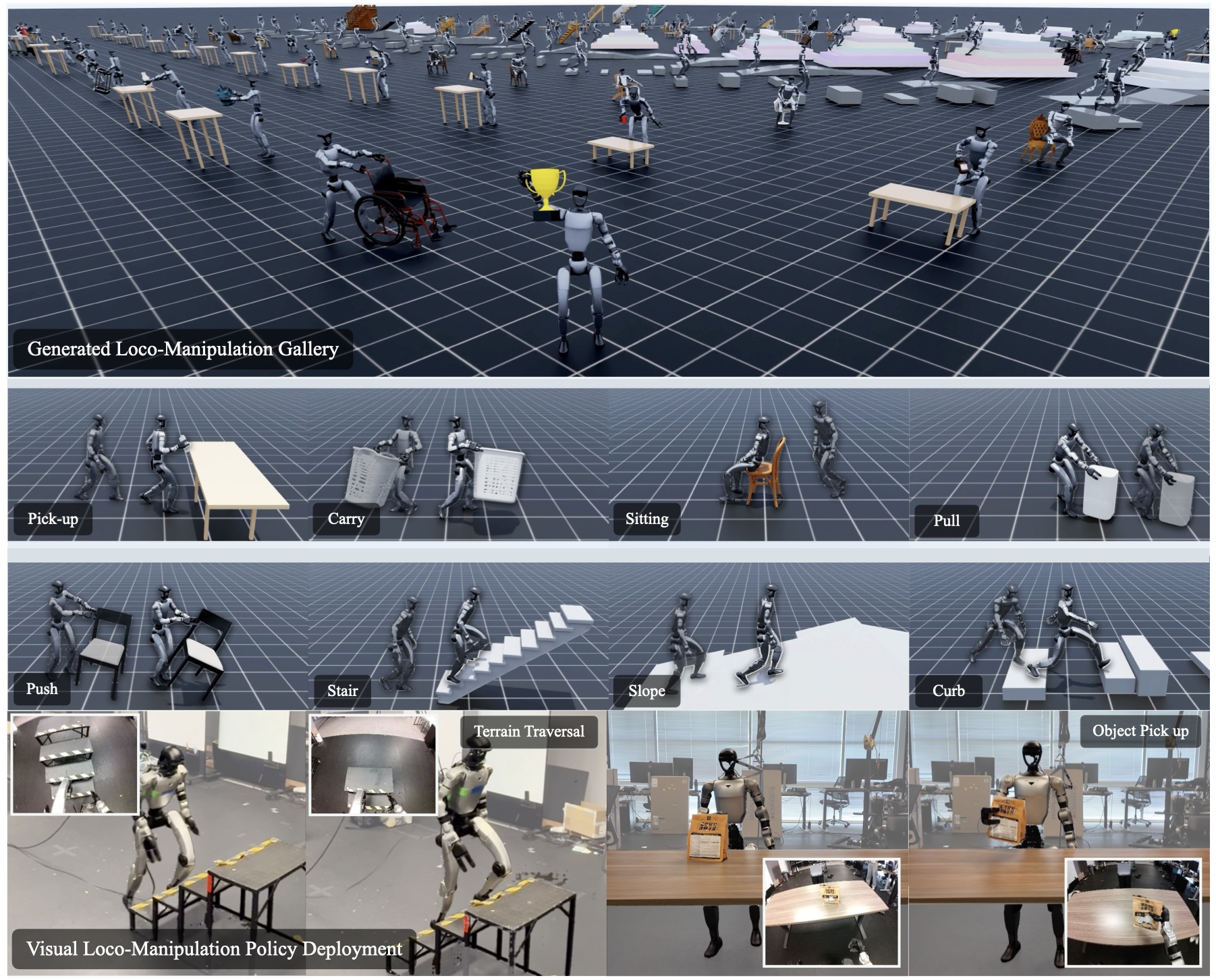}
    \caption{\textbf{From Fully Digital Data Generation to Real-World Deployment.} \METHOD generates humanoid loco-manipulation data by grounding video world models in fully specified 3D scenes, without physical scene rebuilds or robot teleoperation. \textit{Top:} simulated humanoids execute generated references spanning pick-up, whole-body manipulation, sitting, and terrain traversal. \textit{Bottom:} egocentric visual policies trained only on \METHOD-generated data are deployed on a Unitree G1 for stair-climbing and object pick-up.}
    \label{fig:teaser}
\end{figure}

\newpage
\begin{abstract}
Scaling humanoid loco-manipulation requires robot-compatible demonstrations across diverse objects, whole-body motions, and scene geometries, but teleoperation and motion capture are difficult to scale because each collection depends on physical setups, instrumented actors, and robot operation. We present \METHOD, a digital generation pipeline that remains fully virtual until deployment: it grounds video world models in 3D assets and simulator-ready scenes to synthesize interactions without rebuilding physical environments or teleoperating the robot. Rather than reconstructing unconstrained in-the-wild videos, \METHOD starts from fully specified 3D configurations in which object geometry, camera parameters, metric scale, environment depth, and a robot-proportioned character are known before video generation and reused during reconstruction. This privileged setup better conditions 4D recovery, allowing model-based object tracking, human motion estimation, and interaction-aware optimization to reconstruct metric 4D human-object interaction (HOI) trajectories with reduced depth ambiguity and morphology mismatch. We retarget the recovered motions to a humanoid robot and train complementary task-general trackers: an object-aware latent adaptor for manipulation and a scene-aware tracker for terrain traversal. \METHOD produces over 20,000 sequences spanning pick-up, whole-body manipulation, sitting, and terrain traversal. Using only \METHOD-generated data, we train egocentric visual policies through a sim-to-real pipeline and deploy them on a Unitree G1 humanoid, achieving 84\% real-world success on diverse object pick-up and 90\% success on stair-climbing.
\end{abstract}

\abscontent
\section{Introduction}
\label{sec:intro}
Humanoid loco-manipulation requires policies that coordinate whole-body balance, object contact, and scene-aware locomotion across a broad distribution of objects and terrain geometries. Scaling the corresponding demonstration data is challenging because each trajectory must be both physically plausible and executable by the target robot. Teleoperation~\cite{ben2025homie,aldaco2024aloha,ze2025twist2,ze2025twist,khazatsky2024droid} and motion capture~\cite{lu2025humoto,taheri2020grab} provide high-quality demonstrations, but they are difficult to scale: each new object or terrain layout can require human-operated robot demonstrations, instrumented actors, and physical scene reconfiguration. Reconstructing robot-ready 4D trajectories from in-the-wild videos~\cite{zhang2020phosa,xie2022chore,wang2022reconstructing,petrov2023object,hou2023compositional,kim2023ncho,xie2026cari4d} offers broad visual coverage, but requires inferring camera, scale, object geometry, human shape, contacts, and world-space motion from ambiguous monocular observations. Recent advances in 3D asset generation and video world models suggest an alternative route: instead of recovering the entire 3D interaction from an uncontrolled video, can we first specify the 3D scene and then let a world model roll out diverse interactions within it for humanoid policy learning?

We introduce \METHOD, a humanoid-centric data-generation pipeline that remains fully digital until real-world deployment. It uses video foundation models (VFMs) as world models inside a simulator-ready 3D asset pipeline: rather than reconstructing uncontrolled videos into ambiguous 4D scenes, \METHOD first specifies the object, scene geometry, camera, scale, and robot-proportioned character, lets the VFM roll out the interaction from a rendered view of this scene, then recovers it in 4D within this known metric frame. This design addresses two bottlenecks of prior data sources: it avoids repeated physical collection required by teleoperation and motion capture, and it produces robot-trackable trajectories already aligned with simulation for downstream sim-to-real policy training.

The recovered 4D HOI trajectories are retargeted to a Unitree G1 and converted into task-general tracking policies built on a pretrained whole-body controller~\cite{luo2025sonic}. Rather than fitting one controller per sequence or per object, we pool related trajectories so the trackers cover families of manipulation and scene-interaction behaviors.
This stage uses two complementary specializations: an object-aware latent adaptor that augments the frozen whole-body controller with manipulation by modulating its latent tokens and emitting hand actions, and a scene-aware tracker that fine-tunes the controller together with a height-map encoder for terrain-conditioned whole-body control. With this pipeline, we generate a large-scale dataset of over 20,000 humanoid loco-manipulation sequences spanning pick-up, whole-body manipulation, sitting, and terrain traversal. Using only this generated data, we train egocentric visual policies~\cite{he2025viral} with visual domain randomization and camera alignment as a closed-loop sim-to-real validation; deployed on a Unitree G1, the resulting RGB-based policies perform autonomous loco-manipulation on pick-up and stair-climbing tasks.

In summary, \METHOD makes the following contributions: (i) a fully digital humanoid-centric data-generation framework that uses VFMs as world models inside a fully specified 3D asset pipeline, producing over 20,000 physically plausible loco-manipulation sequences; (ii) an interaction-aware 4D HOI reconstruction stack that exploits known geometry, metric scale, camera parameters, environment depth, and a robot-proportioned character; (iii) complementary task-general trackers for reconstructed 4D HOI, pairing object-aware latent adaptation for manipulation with scene-aware height-map conditioning for terrain traversal and sitting; and (iv) an end-to-end sim-to-real validation of \METHOD-generated data through egocentric visual policies deployed on a Unitree G1, achieving 84\% pick-up success and 90\% stair-climbing success in the real world.

\section{Related Work}
\label{sec:related-work}

\noindent\textbf{Human-Object Interaction Generation and Reconstruction.}
Synthesizing human-object interactions (HOI) requires reasoning about human motion, object affordances, and physical contact. Existing data sources rely on motion capture~\cite{taheri2020grab,bhatnagar22behave,jiang2022chairs,huang2022intercap,zhang2023neuraldome,fan2023arctic,li2023object,zhao2023im,kim2024parahome,lu2025humoto} or RGB-based reconstruction~\cite{zhang2020phosa,xie2022chore,wang2022reconstructing,petrov2023object,hou2023compositional,kim2023ncho,xie2026cari4d}, but remain expensive, category-limited, or underconstrained. Learning-based methods synthesize HOI from affordance, language, or vision-language priors~\cite{li2023task,ye2023affordance,zheng2023cams,zhou2022toch,zhang2023artigrasp,Kulkarni_2024_CVPR,diller2023cghoi,li2024controllable,peng2025hoi,Xu_2023_ICCV,jiang2024scaling,dang2025svimo,zhang2025interactanything,xu2024interdreamer,dwivedi2025interactvlm,li2024genzi,wu2025human}, but physical realism and temporal coherence remain challenging. VFM-based pipelines such as DAViD~\cite{kim2025david}, ZeroHSI~\cite{li2026zerohsi}, and related methods~\cite{lou2025zero} use generated videos as priors for 4D HOI recovery, yet typically leave camera, scale, character morphology, object geometry, or environment structure to be inferred after generation. \METHOD instead specifies the 3D scene before generation and reuses it during reconstruction, yielding robot-compatible 4D HOI trajectories grounded by known metric scale, environment geometry, and a robot-proportioned character, facilitating downstream sim-to-real policy learning.

\vspace{1mm}\noindent\textbf{Human Video as Humanoid Supervision.}
Human video has become an increasingly important supervision source for humanoids: large-scale mining, retargeting, and robotized-video generation provide broad pose-control or pretraining data~\cite{mao2024humanoidx,yang2025xhumanoid}, while VideoMimic~\cite{allshire2025videomimic}, HumanX~\cite{wang2026humanx}, and related systems~\cite{weng2025hdmi,yin2025visualmimic,shi2026egohumanoidunlockinginthewildlocomanipulation,yu2025r2r2r} train interaction policies from third-person, monocular, or egocentric videos. In parallel, physics-based control and whole-body imitation~\cite{peng2018deepmimic,peng2021amp,luo2023perpetual,he2024omnih2o,he2025hover,fu2024humanplus,luo2025sonic}, residual adaptors~\cite{zhao2025resmimic}, and multimodal controllers~\cite{he2026ultra,jiang2026wholebodyvla} show how robot-ready references can be converted into executable policies. The shared bottleneck is data: videos still require recovering metric motion, contacts, object state, and scene geometry, while teleoperation, motion capture, wearable interfaces, and generated robot-video demonstrations~\cite{ze2025twist,ze2025twist2,ben2025homie,khazatsky2024droid,lu2025humoto,taheri2020grab,nai2026humi,patel2025rigvid} remain limited by human effort, retargeting, platform dependence, or morphology mismatch. \METHOD addresses this upstream bottleneck by using video world models for behavioral priors while keeping geometry, scale, camera, environment, and target morphology known, producing robot-compatible 4D references for task-general tracking policies.

\section{Method}
\label{sec:data-generation}

Given a 3D object asset $\mathcal{M}^{\mathcal{O}}$, \METHOD produces humanoid loco-manipulation demonstrations comprising humanoid kinematic motion $\{\bm{\Theta}^{\mathcal{R}}_t\}_{t=1}^T$, object kinematic motion $\{\bm{\Theta}^{\mathcal{O}}_t\}_{t=1}^T$, and robot actions $\{\bm{a}^{\mathcal{R}}_t\}_{t=1}^T$. Our data generation pipeline proceeds in three stages. First, we assemble a fully specified 3D configuration with a character prefitted to the target robot, render an initial frame, and feed it into a VFM to synthesize a reference HOI video $\{I_t\}_{t=1}^{T}$ (Sec.~\ref{subsec:hoi-video-gen}). Second, leveraging the known 3D configuration, we reconstruct coherent 4D HOI trajectories $\{(\widetilde{\bm{\Theta}}^{\mathcal{H}}_t, \widetilde{\bm{\Theta}}^{\mathcal{O}}_t)\}_{t=1}^T$ through human pose estimation, object tracking, and joint optimization (Sec.~\ref{subsec:4d-hoi-recon}). Third, we retarget the reconstructed motions to the target humanoid and train task-general tracking policies across each task family (Sec.~\ref{subsec:physics-tracking}). Using the generated data, we further train egocentric visual policies through a sim-to-real pipeline and deploy them on a real Unitree G1 for pick-up and stair-climbing (Sec.~\ref{subsec:visual-policy}).

\subsection{Robot-Centric Human Video Generation}
\label{subsec:hoi-video-gen}
Although one could generate robot videos directly, current VFMs have stronger priors over human manipulation, and human body and hand reconstruction tools are more mature than robot reconstruction tools. We therefore synthesize human interaction videos using a character asset prefitted to the target humanoid, which facilitates retargeting the recovered motion to the robot.
To assemble the 3D configuration, we construct candidate environments using Infinigen~\cite{raistrick2023infinite} and position the human asset in a rest pose alongside the object. We use rigid body simulation~\cite{macklin2016xpbd} to settle the object into a stable, collision-free initial configuration $\bm{\Theta}^{\mathcal{O}}_1$. We then render the first frame using Blender with known camera intrinsics $\bm{C}_K \in \mathbb{R}^{3\times3}$ and extrinsics $\bm{C}_E = (\bm{r}^\mathcal{C}, \bm{t}^\mathcal{C})$. The generated environment serves two purposes: realistic visual context for VFM generation and a ground-truth point cloud for metric-scale depth alignment during reconstruction (Sec.~\ref{subsec:4d-hoi-recon}). A VLM~\cite{openai2024chatgpt} generates an interaction prompt from the rendered frame, and a VFM (\eg, Kling~\cite{klingai2025}), acting as a world model, then rolls out the reference HOI video $\{I_t\}^{T}_{t=1}$ under a static-camera setting that preserves the known camera parameters $(\bm{C}_K, \bm{C}_E)$ for reconstruction.

\begin{figure*}[!t]
    \centering
    \includegraphics[width=1.0\textwidth]{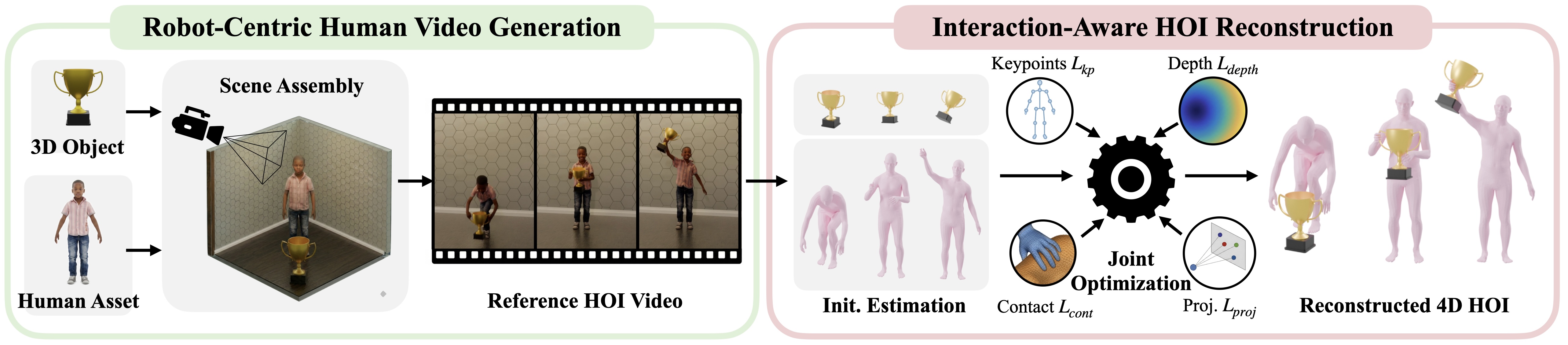}
    \caption{
        \textbf{Asset-Conditioned 4D HOI Generation.} Given a 3D object asset, we render a fully specified 3D scene with a character prefitted to the target humanoid and known camera parameters, synthesize a static-camera interaction video via a VFM conditioned on the rendered frame, and reconstruct metric 4D human-object motion by jointly refining initial human and object trajectory estimates with keypoint, depth, and contact losses anchored to the privileged 3D configuration.}
    \label{fig:pipeline}
\end{figure*}

\subsection{Interaction-Aware HOI Reconstruction}
\label{subsec:4d-hoi-recon}
Given the generated interaction video, we recover the 4D HOI trajectory $\{(\widetilde{\bm{\Theta}}^{\mathcal{H}}_t,\widetilde{\bm{\Theta}}^{\mathcal{O}}_t)\}_{t=1}^T$ in two steps: independent initial estimation of human and object motion, followed by interaction-aware joint optimization that anchors the trajectories to the privileged 3D configuration.

\subsubsection{Initial Motion Estimation}
\label{subsubsec:initial-motion-estimation}
We first estimate human and object motion independently, yielding initial world-space trajectories $\{\widehat{\bm{\Theta}}^{\mathcal{H}}_t\}_{t=1}^T$ and $\{\widehat{\bm{\Theta}}^{\mathcal{O}}_t\}_{t=1}^T$.

\vspace{1mm}\noindent\textbf{Human Motion Estimation.}
For the human body, GENMO~\cite{li2025genmo} provides per-frame SMPL-X~\cite{smplx} pose parameters from the generated video in camera space; the body shape is held fixed at the prefitted character morphology from Sec.~\ref{subsec:hoi-video-gen} rather than re-estimated, so GENMO only contributes per-frame pose parameters. The camera-space motion is then transformed into world coordinates using the known camera extrinsics $\bm{C}_E$. For the hands, WiLoR~\cite{potamias2025wilor} refines per-frame MANO~\cite{romero2017embodied} parameters for the left and right hands independently; missing detections due to partial occlusion or detection failure are filled via temporal linear interpolation and smoothed with a Savitzky-Golay filter~\cite{savitzky1964smoothing} to suppress per-frame jitter. The smoothed hand poses are integrated into the SMPL-X body through wrist inverse-kinematic (IK) alignment, preserving the WiLoR-predicted finger configuration.

\vspace{1mm}\noindent\textbf{Object Pose Tracking.}
For the object, we fine-tune FoundationPose~\cite{wen2024foundationpose} on its proposed synthetic dataset for 5 epochs with the depth channels zeroed at both training and inference to adapt to our RGB-only setup; at inference, the 6-DoF tracker is initialized from the known first-frame pose $\bm{\Theta}^{\mathcal{O}}_1$ and propagates the object pose across all frames. FoundationPose requires known object geometry, texture, and camera parameters, all available in our pipeline by construction, which ensures accurate object tracking. We additionally validate tracking quality by comparing predicted poses against SAM2~\cite{ravi2024sam} segmentation masks and discard sequences with inconsistent geometry (Sec.~\ref{supp:failure-filtering}).

\subsubsection{Joint Optimization}
\label{subsubsec:joint-optimization}
Directly combining the independent reconstructions often produces misaligned interactions (floating contacts, penetration, and depth-scale drift). We therefore jointly refine both trajectories through a global optimization over all frames, holding the hand poses fixed for stability. Rather than optimizing full trajectories directly, we optimize residual motion parameters $\{\Delta\bm{\Theta}^{\mathcal{H}}_t\}_{t=1}^T$ and $\{\Delta\bm{\Theta}^{\mathcal{O}}_t\}_{t=1}^T$; the final poses are $\bm{\Theta}^{\mathcal{H}}_t = \widehat{\bm{\Theta}}^{\mathcal{H}}_t \oplus \Delta\bm{\Theta}^{\mathcal{H}}_t$ and $\bm{\Theta}^{\mathcal{O}}_t = \widehat{\bm{\Theta}}^{\mathcal{O}}_t \oplus \Delta\bm{\Theta}^{\mathcal{O}}_t$, with $\oplus$ denoting residual translation and rotation updates and the 6D rotation representation~\cite{zhou2019continuity} used for continuous parameterization. The full refinement objective is:
\begin{equation}
    L = \lambda_{\text{kp}} L_{\text{kp}} + \lambda_{\text{proj}} L_{\text{proj}} + \lambda_{\text{depth}} L_{\text{depth}} + \lambda_{\text{cont}} L_{\text{cont}} + \lambda_{\text{reg}} L_{\text{reg}},
\end{equation}
yielding the optimized trajectories $\{(\widetilde{\bm{\Theta}}^{\mathcal{H}}_t,\widetilde{\bm{\Theta}}^{\mathcal{O}}_t)\}_{t=1}^T$.

\vspace{1mm}\noindent\textbf{Keypoint Alignment.}
To keep the optimized human trajectory aligned with the generated video, we minimize the distance between projected and detected 2D body and hand keypoints:
\begin{equation}
    L_{\text{kp}} = \frac{1}{T}\sum_{t=1}^{T}\left\|\mathcal{K}^{\mathcal{H}}(\bm{\Theta}^{\mathcal{H}}_t)-p_t\right\|,
\end{equation}
where $p_t \in \mathbb{R}^{J\times3}$ are 2D keypoints obtained from body and hand keypoint estimators~\cite{xu2022vitpose,potamias2025wilor}, and $\mathcal{K}^{\mathcal{H}}(\cdot)$ projects the SMPL-X parameters using the known camera.

\vspace{1mm}\noindent\textbf{Object Projection Alignment.}
Since FoundationPose provides image-aligned object poses, we regularize the optimized object pose to preserve that alignment:
\begin{equation}
    L_{\text{proj}} = \sum_{t=1}^{T}\left\|\mathcal{P}(V^{\mathcal{O}}_t)-\mathcal{P}(\widehat{V}^{\mathcal{O}}_t)\right\|,
\end{equation}
where $\mathcal{P}(\cdot)$ is the camera projection function, and $V^{\mathcal{O}}_t$ and $\widehat{V}^{\mathcal{O}}_t$ are object vertices under the optimized and initial poses.

\vspace{1mm}\noindent\textbf{Depth Alignment.}
Leveraging the known 3D configuration, we first estimate a depth map with MoGe-2~\cite{wang2025moge} and align it to the ground-truth background depth rendered from the environment, recovering metric-scale depth. We then segment human and object regions with SAM2~\cite{ravi2024sam} and unproject them into per-frame point clouds $\mathbf{P}^{\mathcal{H}}_t$ and $\mathbf{P}^{\mathcal{O}}_t$. The depth-alignment loss encourages the reconstructed meshes to match these point clouds:
\begin{equation}
    L_{\text{depth}} = \frac{1}{T}\sum_{t=1}^{T}\mathcal{CD}(V^{\mathcal{H},\mathrm{vis}}_t,\mathbf{P}^{\mathcal{H}}_t)+\mathcal{CD}(V^{\mathcal{O},\mathrm{vis}}_t,\mathbf{P}^{\mathcal{O}}_t),
\end{equation}
where $V^{\mathcal{H},\mathrm{vis}}_t$ and $V^{\mathcal{O},\mathrm{vis}}_t$ are visible mesh vertices and $\mathcal{CD}$ is bidirectional Chamfer distance.

\vspace{1mm}\noindent\textbf{Contact Alignment.}
To encourage physically plausible contact, we query a VLM~\cite{openai2024chatgpt} on uniformly sampled video frames to predict per-frame contact labels (\eg, left or right hand) and propagate each label to its surrounding interval. Using these labels, we identify the relevant SMPL-X vertices $V^{\mathcal{H},\mathrm{cont}}_t$ via SMPL-X part segmentation and apply the contact loss only to frames where contact is detected. Since the image-space losses already enforce projection consistency, the contact loss only needs to resolve depth discrepancies, so we restrict it to object vertices whose projected positions overlap with the contact body region and penalize only their depth offset:
\begin{equation}
    L_{\text{cont}} = \frac{1}{|\mathcal{T}_c|}\sum_{t\in\mathcal{T}_c}\mathcal{CD}_z(V^{\mathcal{H},\mathrm{cont}}_t,V^{\mathcal{O},\mathrm{cont}}_t), \quad V^{\mathcal{O},\mathrm{cont}}_t=\mathcal{F}(V^{\mathcal{O}}_t,V^{\mathcal{H},\mathrm{cont}}_t),
\end{equation}
where $\mathcal{T}_c$ is the set of frames where contact is detected. The filter $\mathcal{F}$ projects both vertex sets to screen space and keeps the object vertices whose projections fall within a distance threshold of the contact body vertices, and $\mathcal{CD}_z(\cdot,\cdot)$ is a depth-only bidirectional Chamfer distance that penalizes the positional difference along the viewing direction. $L_{\text{cont}}$ is disabled for terrain-only sequences without hand-object interaction.

\vspace{1mm}\noindent\textbf{Regularization.}
The regularization term decomposes as $L_\text{reg} = L_\text{foot} + L_\text{vel} + L_\text{smooth}$. $L_\text{foot}$ leverages per-frame foot contact labels from GENMO~\cite{li2025genmo} to penalize foot vertex displacement during detected contact frames, suppressing foot skating. $L_\text{vel}$ regularizes the optimized pelvis velocity to match GENMO's global-space velocity estimate, suppressing the depth-direction oscillations that camera-space estimates exhibit under depth-scale ambiguity. $L_\text{smooth}$ penalizes the first- and second-order temporal finite differences of the human and object mesh vertex positions for temporal coherence.

\subsection{Task-General Loco-Manipulation Tracking}
\label{subsec:physics-tracking}
\begin{figure*}[!t]
    \centering
    \includegraphics[width=1.0\textwidth]{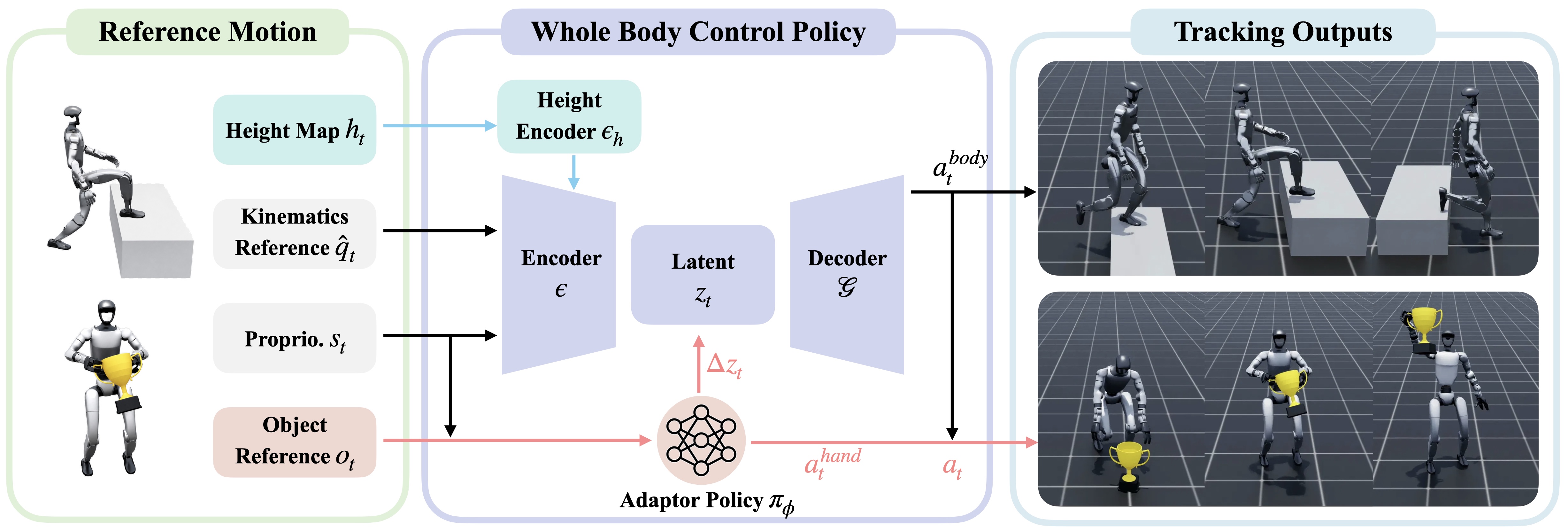}
    \caption{\textbf{Task-General Tracking via Complementary Controller Adaptation.} Retargeted 4D HOI trajectories are converted into robot-action data by adapting different parts of a pretrained whole-body controller. Object-manipulation trajectories are used to train an object-aware adaptor $\pi_\phi$ while the controller remains frozen: the adaptor observes proprioception $\bm{s}_t$ and an object reference $\bm{o}_t$, injects a latent residual $\Delta\bm{z}_t$, and emits hand actions $\bm{a}^{\mathrm{hand}}_t$. Terrain traversal and chair-sitting trajectories are used to fine-tune the controller with a scene-aware height encoder $\epsilon_h$: the encoder maps the local height map $h_t$ into terrain context for scene-conditioned whole-body control.}
    \label{fig:motion-tracking}
\end{figure*}

This robot-proportioned reconstruction allows GMR~\cite{ze2025gmr} to retarget the SMPL-X motion $\{\widetilde{\bm{\Theta}}^{\mathcal{H}}_t\}_{t=1}^T$ to the Unitree G1 with reduced morphology mismatch, better preserving hand-object and body-scene contacts. The result is a kinematic reference motion $\{\widetilde{\bm{q}}_t\}_{t=1}^T$ in the robot's joint space, while the reconstructed object trajectory $\{\widetilde{\bm{\Theta}}^{\mathcal{O}}_t\}_{t=1}^T$ provides the reference object pose. We then train tracking policies built on SONIC~\cite{luo2025sonic}, a pretrained whole-body controller, to convert these retargeted 4D HOI trajectories into robot-action data. Rather than fitting a controller per sequence or per object, we train task-general policies across each task family; as related trajectories are added, existing policies provide initialization for fine-tuning, amortizing adaptation across the pool. As outlined in Fig.~\ref{fig:motion-tracking}, we instantiate this stage with two complementary specializations: an \emph{object-aware} latent adaptor trained on object-manipulation trajectories and a \emph{scene-aware} tracker trained on terrain traversal and chair-sitting trajectories. The object-aware adaptor adds hand actions and modulates the latent tokens fed to the controller's frozen action decoder, enabling manipulation while preserving the locomotion prior; the scene-aware tracker fine-tunes the controller with a height-map encoder, improving terrain-conditioned whole-body control for traversal and scene interaction.

\vspace{1mm}\noindent\textbf{Object-Aware Tracking.}
For object-manipulation 4D HOI trajectories, we extend the pretrained whole-body controller with an object-aware adaptor policy $\pi_\phi$ that modulates its latent token space and emits hand actions, giving the frozen controller object-manipulation capability while preserving its pretrained locomotion behavior. The controller encodes kinematic motion targets into a discrete latent token $\bm{z}_t = \mathcal{E}(\widetilde{\bm{q}}_t)$ via finite scalar quantization and decodes them into joint-level actions through $\mathcal{G}(\bm{z}_t)$. We keep its encoder, quantizer, and decoder frozen, then train only $\pi_\phi$ to inject manipulation-specific residuals. The adaptor observes proprioception $\bm{s}_t$ and an object reference $\bm{o}_t$ consisting of the object pose in robot body frame, hand-to-object transforms, finger contact forces, a precomputed basis point set (BPS)~\cite{prokudin2019efficient} shape encoding, and, critically, delta observations encoding the difference between the reference future object pose and the current simulated pose. It outputs a 64-dim latent residual $\Delta\bm{z}_t$ together with a 2-dim binary hand primitive $\bm{a}^{\mathrm{hand}}_t$ that maps to 7 finger DoFs per hand:
\begin{equation}
    (\Delta\bm{z}_t,\bm{a}^{\mathrm{hand}}_t)=\pi_\phi(\bm{s}_t,\bm{o}_t),
    \quad
    \bm{a}^{\mathrm{body}}_t=\mathcal{G}(\bm{z}_t+\lambda\Delta\bm{z}_t),
\end{equation}
with $\lambda = 0.1$ scaling the residual before FSQ quantization. Each hand primitive produces a binary open/close grasp signal, which is mapped to 7 finger joint positions per hand via predefined grasp configurations. The BPS encoding provides the adaptor with object-shape awareness, enabling a single $\pi_\phi$ to track motions across diverse object geometries. An auxiliary $\ell_2$ penalty is applied on $\Delta\bm{z}_t$ to encourage the adapted latent to remain close to the pretrained controller's behavior.


\vspace{1mm}\noindent\textbf{Scene-Aware Tracking.}
For tasks involving scene-aware interactions, such as stepping over curbs, climbing up stairs, and sitting on chairs, the controller's flat-ground prior is not directly usable. We therefore fine-tune the controller on a mixture of reconstructed 4D HOI scene-interaction trajectories and its original flat-ground data, while augmenting the encoder input with a local height map $h_t$ around the robot processed by a 2D-convolutional projector $\epsilon_h$. This preserves the base locomotion distribution while teaching the controller to adapt whole-body motions to terrain and scene geometry. To stabilize the training, in addition to the action decoder $\mathcal{G}$, we train a parallel kinematic decoder $\mathcal{G}_{\mathrm{rec}}$ that reconstructs the input motion targets to provide an auxiliary MSE loss that regularizes the latent to remain faithful to the trajectory.

\vspace{1mm}\noindent\textbf{Reward Design.}
Both trackers share a \textit{motion-tracking reward} $R^{\mathrm{motion}}_t$ that encourages the simulated robot to follow the retargeted reference, together with regularization penalties $R^{\mathrm{reg}}_t$ (\eg, action rate and joint limits) for smoothness and safety. The motion-tracking reward is a sum of exponential terms over reference--simulation discrepancies:
\begin{equation}
    R^{\mathrm{motion}}_t = \sum_{i} w_i \exp\!\left(-\frac{\|\widetilde{\bm{x}}_{i,t} - \bm{x}_{i,t}\|^2}{\sigma_i^2}\right),
\end{equation}
where $\widetilde{\bm{x}}_{i,t}$ and $\bm{x}_{i,t}$ are reference and simulated quantities spanning root pose, per-body positions and orientations, and linear and angular velocities; for object-aware tracking we additionally boost the weight on the wrist links to encourage accurate hand placement.
For object-aware tracking, the total reward adds an object term and a contact-gated grasp term, $R_t = R^{\mathrm{motion}}_t + R^{\mathrm{reg}}_t + R^{\mathrm{obj}}_t + \mathbb{1}\{C_t\}\,R^{\mathrm{grasp}}_t$, where $\mathbb{1}\{C_t\}$ is a per-frame contact indicator carried by the reconstructed trajectory, so the grasp reward is active only during contact phases. The \textit{object tracking reward} penalizes deviation from the reference object pose:
\begin{equation}
    R^{\mathrm{obj}}_t = w_{p}\exp\!\left(-\alpha_p \|\widetilde{\bm{p}}^{\mathcal{O}}_t - \bm{p}^{\mathcal{O}}_t\|\right) + w_{r}\exp\!\left(-\alpha_r \|\widetilde{\bm{r}}^{\mathcal{O}}_t \ominus \bm{r}^{\mathcal{O}}_t\|\right),
\end{equation}
with scaling coefficients $\alpha_p, \alpha_r > 0$. The \textit{grasp reward} combines three terms per hand:
\begin{align}
    R^{\mathrm{grasp}}_t = \;& w_c \underbrace{\min\!\left(\frac{N^{\mathrm{contact}}_t}{N_{\min}},\; 1\right)}_{\text{contact time}} + w_d \underbrace{\left[-\cos(\bm{d}^{\mathrm{thumb}}_t,\, \bm{d}^{\mathrm{index}}_t)\right]^{+}}_{\text{grasp pose}} \nonumber \\
    &+ \; w_f \underbrace{\exp\!\left(-\gamma\, \tfrac{1}{N_f}\textstyle\sum_{j} \|\bm{f}_{j,t} - \bm{c}_t\|\right)}_{\text{contact proximity}}.
\end{align}
The first term rewards sustained finger contact with the object (saturating at $N_{\min}$ contacts), the second encourages the thumb and index finger to approach from opposing sides for a stable pinch grasp ($\bm{d}^{\mathrm{thumb}}_t$, $\bm{d}^{\mathrm{index}}_t$ are vectors from the object center to each fingertip), and the third draws all fingertips $\bm{f}_{j,t}$ toward the object contact centroid $\bm{c}_t$. Since scene-aware tasks involve no hand-object interaction, the scene-aware tracker uses only $R^{\mathrm{motion}}_t$ and $R^{\mathrm{reg}}_t$.

\vspace{1mm}\noindent\textbf{Training.}
We train each stage with PPO~\cite{schulman2017proximal} in Isaac Lab on 64 NVIDIA L40 GPUs, running for 30{,}000 iterations with 1{,}024 environments per GPU. Object-aware tracking updates only $\pi_\phi$ while the controller's encoder, quantizer, and decoder remain frozen. Scene-aware tracking instead fine-tunes the controller together with the height-map encoder $\epsilon_h$. Multiple reference motions are trained jointly within each task family, with environments sampling motions from a shared 4D HOI pool, and we apply reference state initialization at every episode reset.

Full reward definitions and implementation details are provided in Appendix~\ref{supp:tracking-details}.

\subsection{Sim-to-Real Deployment}
\label{subsec:visual-policy}
For sim-to-real deployment, we distill the object-aware and scene-aware tracking policies into separate egocentric visual policies~\cite{luo2025sonic,he2025viral,chi2023diffusionpolicy} for object pick-up and stair-climbing, respectively. The deployed models consume head-camera RGB inputs and output the latent tokens of the SONIC controller, and are trained with domain randomization to facilitate sim-to-real transfer. To deploy on the real Unitree G1, we connect the robot to a desktop with an NVIDIA RTX 5090 GPU and stream visual and proprioceptive input to the desktop before streaming robot actions to the G1. We use a Luxonis OAK-D W camera on the G1 and run inference at 10\,Hz.

\section{Results}
\begin{figure}[!p]
    \centering
    \includegraphics[width=0.93\textwidth]{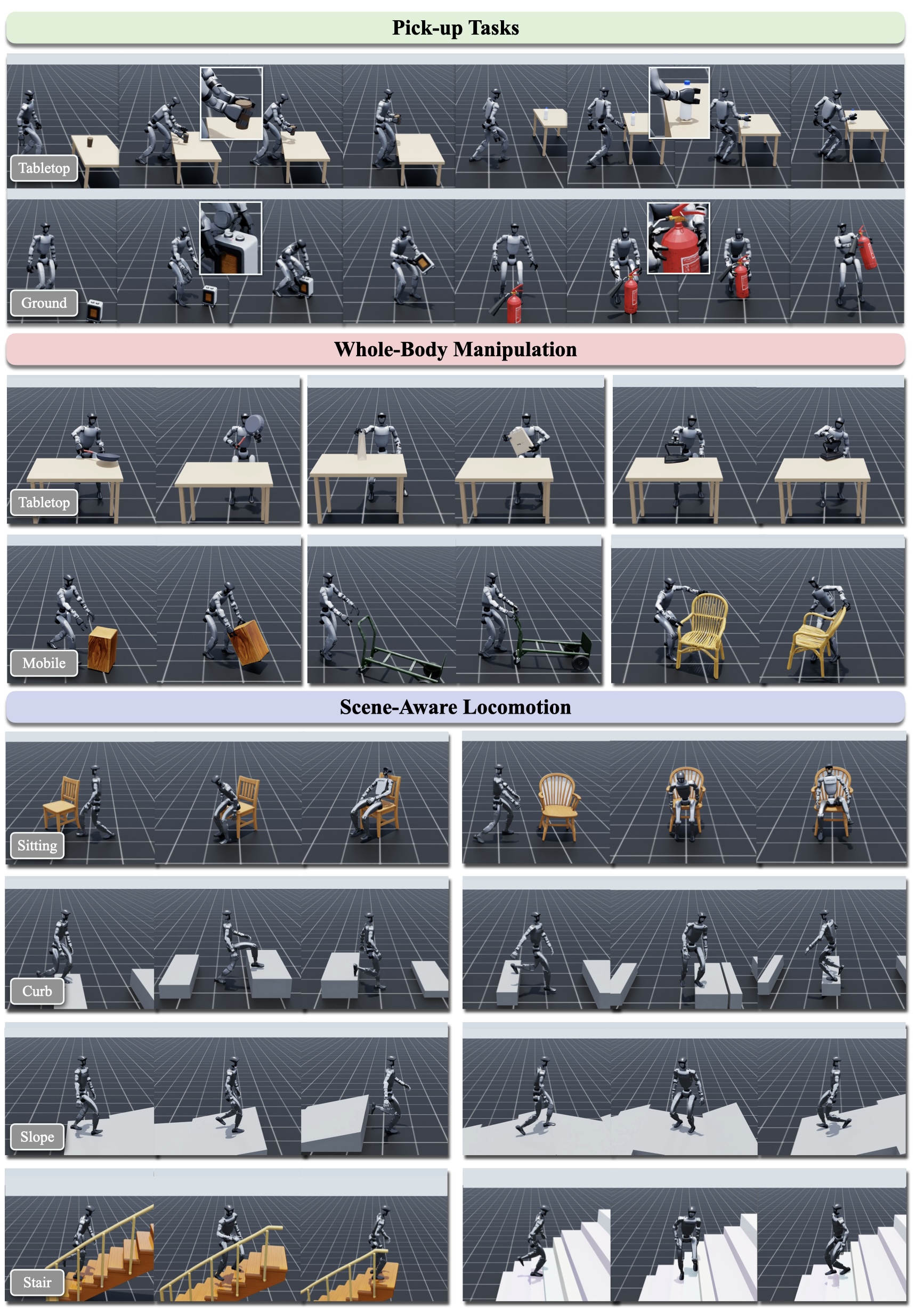}
    \caption{
        \textbf{Generated Loco-Manipulation Data.} Representative simulated Unitree G1 executions from the generated dataset span pick-up, whole-body manipulation, sitting, and terrain traversal across diverse objects and scene geometries.
    }
    \label{fig:loco-manipulation-data}
    \vspace{-2mm}
\end{figure}

Our experiments cover the three stages of the \METHOD pipeline. We first evaluate whether the generated 4D HOI sequences are more physically executable than existing generation baselines. We then ask whether these 4D HOI sequences can be converted into task-general loco-manipulation policies at scale, rather than only replayed through per-sequence tracking. Finally, we demonstrate the practical value of the generated data by deploying egocentric visual policies on the real robot for autonomous loco-manipulation.

\subsection{Human-Object Interaction Generation}
\label{subsec:hoi-generation-results}

\noindent\textbf{Setup.}
We compare the HOI generation component of \METHOD against training-based (CHOIS~\cite{li2024controllable}, HOIDiff~\cite{peng2025hoi}) and training-free (DAViD~\cite{kim2025david}) 4D HOI generation approaches on a shared evaluation set of 20 everyday objects from ComAsset~\cite{kim2024beyond}. Detailed baseline configurations are provided in Appendix~\ref{supp:additional-results}.

\vspace{1mm}\noindent\textbf{Metrics.}
We evaluate the generated 4D HOI sequences along three axes. (i) \textit{Geometric quality:} \emph{contact distance} (Contact) is the average top-$k$ vertex-to-vertex distance between the SMPL-X human surface and the object surface; \emph{penetration ratio} (Pen.) is the percentage of SMPL-X vertices that interpenetrate the object mesh. (ii) \textit{Perceptual realism:} \emph{Interaction Score} (Inter.\ Score) is a VLM rating~\cite{openai2024chatgpt} of sampled keyframes on a 1-5 scale based on physical plausibility and affordance correctness; \emph{motion smoothness} for the human (Human Smo.) and the object (Obj Smo.) is the second-order temporal derivative of their vertex trajectories. (iii) \textit{Physical executability:} we apply InterMimic~\cite{xu2025intermimic}, an SMPL-X humanoid tracking framework, to reproduce each method's 4D HOI sequences in physics simulation using humanoids built from capsule primitives that conform to the input body shape (no motion retargeting required). We report the full-body mean per-joint position deviation (Body Dev.) and the mean object surface deviation (Obj Dev.), and define the tracking success rate (SR) as the fraction of frames where the normalized full-body and object deviations (divided by the object's maximum dimension) are both below $0.25$.

\begin{table}[!t]
    \centering
    \footnotesize
    \setlength\tabcolsep{3pt}
    \resizebox{\textwidth}{!}{%
    \begin{tabular}{lccccccccc}
        \toprule
        & \multicolumn{2}{c}{\textbf{Geometric}} & \textbf{Perceptual} & \multicolumn{2}{c}{\textbf{Smoothness}} & \multicolumn{3}{c}{\textbf{Physical Executability}} \\
        \cmidrule(lr){2-3} \cmidrule(lr){4-4} \cmidrule(lr){5-6} \cmidrule(lr){7-9}
        \textbf{Methods}
        & Contact $\downarrow$ & Pen. $\downarrow$ & Inter. Score $\uparrow$
        & Human Smo. $\downarrow$ & Obj Smo. $\downarrow$
        & SR $\uparrow$ & Body Dev. $\downarrow$ & Obj Dev. $\downarrow$ \\ \midrule
        HOIDiff~\cite{peng2025hoi}      & 0.012          & 2.07\%          & 1.79          & 0.0043          & 0.0118          & 15.8\%          & 0.2120          & 0.3352 \\
        CHOIS~\cite{li2024controllable} & 0.034          & 3.74\%          & 2.47          & 0.0055          & 0.0062          & 10.5\%          & 0.2564          & 0.3642 \\
        DAViD~\cite{kim2025david}       & 0.246          & 1.46\%          & 2.74          & \textbf{0.0024} & 0.0605          & 24.0\%          & 0.4723          & 0.5826 \\
        \midrule
        Ours                            & \textbf{0.008} & \textbf{0.90\%} & \textbf{3.58} & 0.0033          & \textbf{0.0022} & \textbf{88.9\%} & \textbf{0.0913} & \textbf{0.0851} \\
        \bottomrule
    \end{tabular}%
    }
    \caption{\textbf{Comparison of HOI Generation.} Geometric quality, perceptual realism (Interaction Score, 1-5 scale), motion smoothness, and physics-based tracking on the shared 20-object evaluation set.}
    \label{tab:hoi-comparison}
\end{table}

\begin{table}[!t]
    \centering
    \setlength\tabcolsep{6pt}
    \small
    \begin{tabular}{lccc}
        \toprule
        \textbf{Method} & SR $\uparrow$ & ObjPos $\downarrow$ & MPJPE-L $\downarrow$ \\ \midrule
        HDMI~\cite{weng2025hdmi}        & 48.5\% & 0.283 & 122.3 \\
        ResMimic~\cite{zhao2025resmimic} & 49.2\% & 0.393 & 80.9 \\ \midrule
        Ours w/o SONIC          & 45.0\%          & 0.395          & 243.5 \\
        Ours w/o $\pi_\phi$     & 39.7\%          & 0.303          & \textbf{37.1} \\
        Ours w/o Rel. Obs.     & 57.9\%          & 0.257          & 43.0 \\
        \midrule
        Ours (Full)             & \textbf{81.4\%} & \textbf{0.135} & 41.8 \\
        \bottomrule
    \end{tabular}
    \caption{\textbf{Task-General Loco-Manipulation Tracking.} Comparison against loco-manipulation baselines (top) and ablations of \METHOD's object-aware adaptor (bottom). Metrics: success rate (SR), object position error (ObjPos), and local per-joint error (MPJPE-L).}
    \label{tab:ablation-rl}
\end{table}

\vspace{1mm}\noindent\textbf{Comparison.}
As shown in Table~\ref{tab:hoi-comparison}, \METHOD achieves the strongest performance across nearly all metrics: the lowest contact distance and penetration ratio, the highest interaction score, the smoothest object trajectories, and, by a large margin, the highest tracking success rate with the lowest body and object deviation. This confirms that the generated 4D HOI sequences are both perceptually realistic and physically executable, making them well-suited for downstream robot learning. We further conduct a user study and present additional qualitative comparisons in Appendix~\ref{supp:additional-results}.


\begin{figure}[!t]
    \centering
    \includegraphics[width=0.98\textwidth]{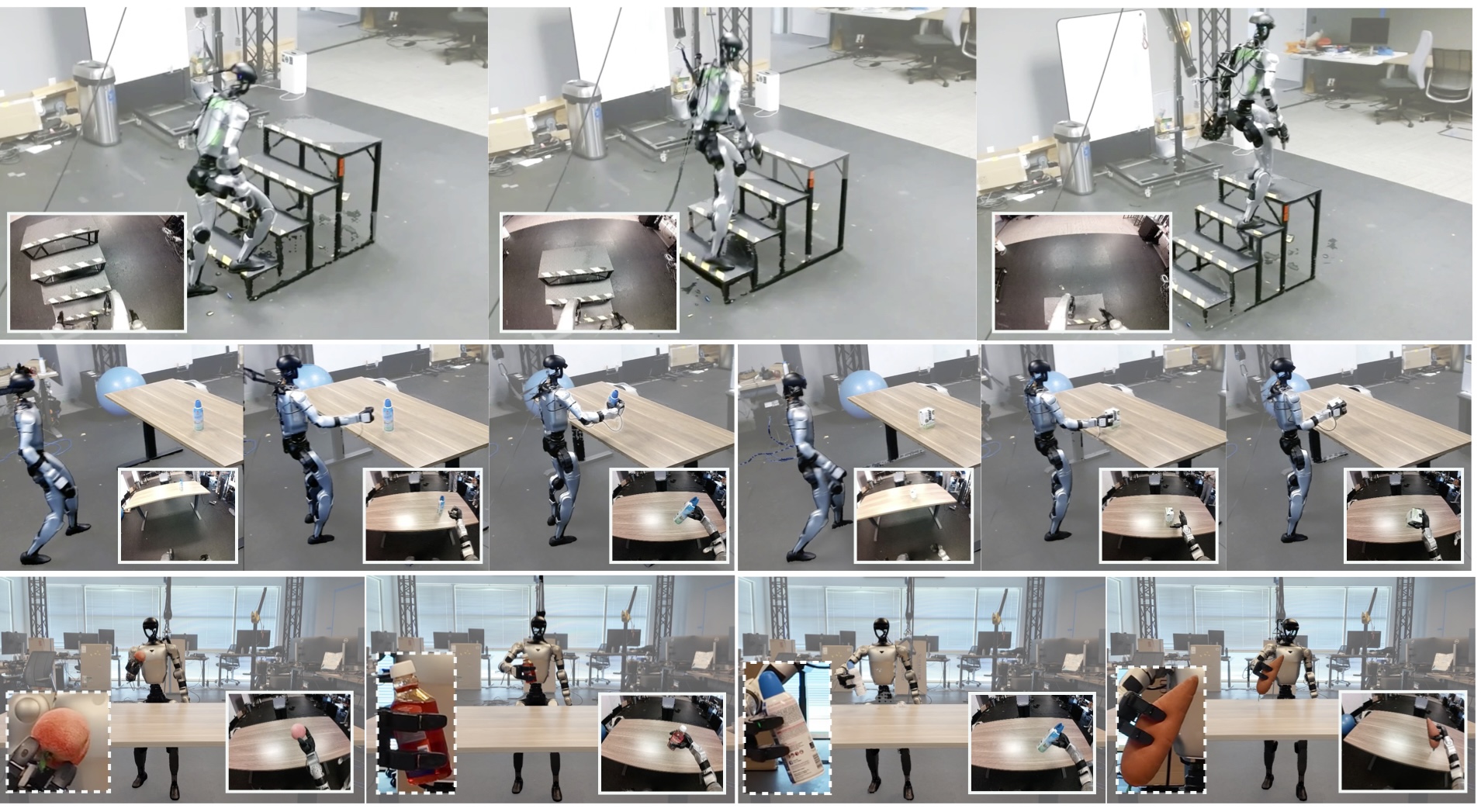}
    \caption{
        \textbf{Sim-to-Real Deployment.} Egocentric visual policies trained only on \METHOD-generated data transfer to a Unitree G1 for object pick-up and stair-climbing.
    }
    \label{fig:sim-to-real-deployment}
\end{figure}

\begin{table}[!t]
    \centering
    \setlength\tabcolsep{2pt}
    \renewcommand{\arraystretch}{1.05}
    \scriptsize
    \resizebox{\columnwidth}{!}{%
    \begin{tabular}{lcccccccccccc}
        \toprule
        & \multicolumn{6}{c}{\textbf{Seen Objects}}
        & \multicolumn{6}{c}{\textbf{Unseen Objects}} \\
        \cmidrule(lr){2-7} \cmidrule(lr){8-13}
        & Cube & Apple & Tea Box & Carrot & Wet Wipes & \textbf{Avg.}
        & Spray Can & Lint Roller & Peach & Flashlight & Medicine Bottle & \textbf{Avg.} \\
        \midrule
        SR $\uparrow$
        & 100\% & 60\% & 100\% & 70\% & 90\% & \textbf{84\%}
        & 100\% & 50\% & 90\% & 80\% & 80\% & \textbf{80\%} \\
        \bottomrule
    \end{tabular}%
    }
    \caption{\textbf{Sim-to-Real Pick-up Results for Seen and Unseen Objects.} Success rates are computed over 10 trials.}
    \label{tab:object_success_rates}
\end{table}

\subsection{Task-General Loco-Manipulation Tracking}
\label{subsec:task-general-tracking-results}

\noindent\textbf{Scaling Loco-Manipulation Data.}
Using the asset-conditioned generation pipeline, we generate a large-scale loco-manipulation dataset for the Unitree G1 with 1{,}000 object assets sourced from Robocasa~\cite{nasiriany2024robocasa}, ComAsset~\cite{kim2024beyond}, OMOMO~\cite{li2023object}, and Hunyuan3D~\cite{hunyuan3d2025hunyuan3d}, paired with 1{,}000 procedurally generated terrain configurations. The resulting dataset contains over 20,000 sequences (Fig.~\ref{fig:loco-manipulation-data}) spanning four categories. \textit{Pick-up} covers tabletops and the ground, exercising diverse grasp strategies across varying object shapes and placement heights. \textit{Whole-body manipulation} covers tabletop manipulation motions and mobile interactions in which the robot carries, pushes, or repositions larger items such as boxes and carts while walking. \textit{Sitting} spans diverse chair styles, requiring approach, lower-body adjustment, and settling into a seated posture. \textit{Terrain traversal} covers procedurally generated curbs, slopes, and stairs, a setting essential for real-world deployment but underrepresented in existing datasets.

\vspace{1mm}\noindent\textbf{Baseline Comparison.}
We compare our method against two recent humanoid loco-manipulation baselines, HDMI~\cite{weng2025hdmi} and ResMimic~\cite{zhao2025resmimic}, using their official implementations on a benchmark of 124 motions across 43 objects.
We report manipulation success rate (SR), object position error (ObjPos), and local mean per-joint position error (MPJPE-L); SR is the fraction of episodes where the average object position error falls below 20\,cm.
Both train whole-body tracking policies from human references but differ from our approach in two key ways. First, neither method actuates per-finger DoFs, so their evaluated interactions rely on whole-arm or whole-body contact such as carrying, lifting, and pushing. Second, both train a separate policy per task: ResMimic trains a per-task residual on top of a general motion-tracking base, while HDMI trains one specialist policy per task. In contrast, \METHOD trains task-general policies across large in-family pools of 4D HOI trajectories. As shown in the top block of Table~\ref{tab:ablation-rl}, \METHOD outperforms the baselines by a large margin across success rate, object position error, and body tracking accuracy.

\vspace{1mm}\noindent\textbf{Ablation Study.}
We ablate the object-aware latent adaptor for manipulation cases on the same benchmark and report results in Table~\ref{tab:ablation-rl} (bottom block).
Removing SONIC and training from scratch substantially degrades body tracking and reduces success rate. Disabling the latent adaptor $\pi_\phi$ (i.e., vanilla SONIC) yields the lowest manipulation success rate despite the best body tracking, indicating that accurate body imitation alone is insufficient for object interaction. Replacing relative object observations with absolute ones also decreases success rate.

\subsection{Sim-to-Real Deployment}
\label{subsec:deployment-results}

To demonstrate \METHOD's real-world applicability, we deploy trained egocentric visual policies for stair-climbing and diverse object pick-up. For stair-climbing, a policy trained on diverse terrain-traversal sequences from \METHOD achieves a 90\% real-world success rate, as shown in Fig.~\ref{fig:sim-to-real-deployment}. For object pick-up, we train on 200 approach-and-pick-up sequences per object across cubes, apples, tea boxes, carrots, and wet wipes. The resulting policy achieves an 84\% real-world success rate, as shown in Table~\ref{tab:object_success_rates}, and transfers effectively to unseen objects, attaining an 80\% success rate.

\section{Conclusion}
We presented \METHOD, a fully digital pipeline for generating humanoid loco-manipulation data from video world models grounded in 3D assets, requiring the physical robot and environment only at deployment. Instead of reconstructing uncontrolled videos, \METHOD starts from fully specified 3D scenes where object geometry and texture, camera parameters, metric scale, environment depth, and robot-proportioned morphology are available by construction. This privileged setup turns key ambiguities in 4D HOI reconstruction into controlled inputs, enabling model-based object tracking, metric depth alignment, and interaction-aware optimization to recover robot-compatible trajectories. The recovered motions are retargeted to the Unitree G1 and converted into complementary task-general trackers: object-aware latent adaptation for manipulation and scene-aware height-map conditioning for terrain traversal and sitting. From over 20,000 generated sequences, we train egocentric visual policies using only \METHOD-generated data and deploy them on a real G1, achieving 84\% pick-up success across diverse objects and 90\% stair-climbing success. These results suggest that world-model-generated data, grounded in known 3D scenes, can complement teleoperation and motion capture as a scalable route toward humanoid loco-manipulation.

\section{Limitations}
Our pipeline assumes 3D object assets, simulator-ready scene setup, and a video world model that follows the requested interaction. Reconstruction quality degrades under severe occlusion, fast motion, or inconsistent object appearance from the VFM, and the failure-filtering step discards a non-trivial fraction of sequences. The task-general tracking policies amortize learning over related 4D HOI pools, but still require training or fine-tuning when the motion family changes substantially.

\clearpage

\bibliographystyle{plainnat}
\bibliography{main}

\appendix
\newpage
\clearpage

\section{Human-Object Interaction Generation Details}

\subsection{Video Generation}
We construct two candidate scene configurations using Infinigen~\cite{raistrick2023infinite}: an indoor floor-only environment and a furnished room with a table (Fig.~\ref{fig:scene_config}). For each input object, a VLM~\cite{openai2024chatgpt} determines whether it should be placed on the floor (\eg, a sofa) or on the table (\eg, a frypan) based on its typical affordances. We pair the object with a human asset prefitted to the Unitree G1's morphology, render the initial frame, and use a VLM~\cite{openai2024chatgpt} to generate a text prompt describing the intended interaction. The rendered frame and prompt are then passed to Kling 2.5 Turbo Pro~\cite{klingai2025}, which supports generating 5 or 10 second videos at 24~fps with resolution of 1920$\times$1080. We can optionally render an ending frame to control the final positions of the human and object, and produce longer sequences auto-regressively by feeding the last frame of each segment as the initial frame of the next.

\afterpage{%
\begin{figure}[t]
    \centering
    \includegraphics[width=1.0\linewidth]{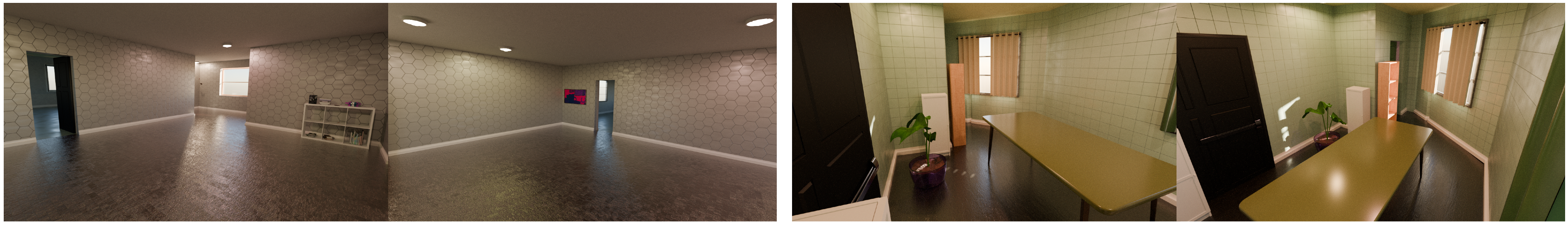}
    \caption{\textbf{Candidate Scene Configurations.} Two pre-built 3D scene templates used to place objects: an indoor floor-only environment for ground-level objects (\eg, a sofa), and a furnished room with a table for tabletop objects (\eg, a frypan). A VLM selects between the two based on each object's typical affordances.}
    \label{fig:scene_config}
\end{figure}%
}

\subsection{Generation Runtime}
\begin{wraptable}{r}{0.45\textwidth}
    \centering
    \vspace{-\baselineskip}
    \setlength\tabcolsep{8pt}
    \small
    \begin{tabular}{lc}
        \toprule
        \textbf{Stage} & \textbf{Time} \\ \midrule
        Video Generation (Kling API)    & $\sim$1 min \\
        Human Motion Estimation         & $\sim$2 min \\
        Object Pose Tracking            & $\sim$1 min \\
        Optimization Preprocessing      & $\sim$2 min \\
        Joint Optimization              & $\sim$8 min \\
        \midrule
        \textbf{Total}                  & $\sim$14 min \\ \bottomrule
    \end{tabular}
    \caption{\textbf{Runtime Breakdown.} Wall-clock time per stage of the \METHOD 4D HOI generation pipeline, measured on a single NVIDIA A100 GPU for a 5-second, 121-frame sequence.}
    \label{tab:runtime}
\end{wraptable}
Table~\ref{tab:runtime} reports the wall-clock time for each stage of the 4D HOI generation pipeline per sequence (a 5-second video at 24~fps, 121 frames), measured on a single NVIDIA A100 GPU. The full pipeline takes approximately 14 minutes per sequence. Video generation and initial motion estimation (human and object) together account for about 4 minutes. The optimization preprocessing stage, which runs MoGe-2~\cite{wang2025moge} for metric depth estimation and SAM2~\cite{ravi2024sam} for human and object segmentation to produce per-frame point clouds as optimization targets, takes roughly 2 minutes. The joint optimization stage dominates at approximately 8 minutes, as it jointly optimizes human and object trajectories across all frames.

\subsection{Failure Case Filtering}
\label{supp:failure-filtering}
While image-to-video models produce realistic HOI sequences, they may introduce artifacts such as texture inconsistencies or geometry mismatches across frames, causing FoundationPose to lose tracking. To automatically filter such failures, we compare SAM2~\cite{ravi2024sam} object masks $\{\widehat{\mathcal{M}}_t\}_{t=1}^T$ against rendered silhouettes $\{\mathcal{M}_t\}_{t=1}^T$ from the predicted poses, and compute the mask tracking error:
\begin{equation}
    e^\mathcal{M} = \sum_{t=1}^T \frac{\text{Sum}\left( (1 - {\mathcal{M}}_t)\cdot \widehat{ \mathcal{M}}_t \right)}{\text{Sum} \left({ \mathcal{M}}_t\right)},
\end{equation}
where $\text{Sum}(\cdot)$ counts non-zero pixels. This measures the fraction of SAM2-tracked mask pixels not covered by the predicted mask. We discard sequences where $e^\mathcal{M}$ exceeds $\tau = 0.2$, effectively removing cases caused by fast motion, blurry frames, or inconsistent object appearance.

\section{Task-General Loco-Manipulation Tracking Details}
\label{supp:tracking-details}

We train two physics-based tracking policies on top of SONIC~\cite{luo2025sonic}, a pretrained whole-body controller: an \emph{object-aware adaptor} for manipulation 4D HOI trajectories and a \emph{scene-aware tracker} for terrain- and chair-conditioned 4D HOI trajectories. Per-tracker policy observations are summarized in Table~\ref{tab:observations} and reward terms in Table~\ref{tab:rewards}.

\subsection{Object-Aware Adaptor}

\noindent\textbf{Architecture.}
The adaptor policy $\pi_\phi$ is a 3-layer MLP with hidden dimensions $[512, 256, 128]$ and SiLU activations. It outputs a 66-dim meta-action: 64 dims for the latent residual (matching the controller's token dim of $2 \times 32$) and 2 dims for left/right hand primitives. The residual is scaled by $\lambda = 0.1$ and added to the controller's encoder output \textit{before} finite scalar quantization, then decoded to 29 body joint position targets. Each hand primitive is passed through a sigmoid and thresholded to a binary open/close signal, mapped to 7 finger joint positions per hand via predefined grasp configurations. The critic is a separate 3-layer MLP $[512, 256, 128]$ that receives privileged observations including full body state and ground-truth contact flags.

\vspace{1mm}\noindent\textbf{Training.}
We train $\pi_\phi$ with PPO~\cite{schulman2017proximal} on 64 NVIDIA L40 GPUs with 1{,}024 parallel environments per GPU in Isaac Lab for 30{,}000 iterations; the pretrained encoder, FSQ quantizer, and decoder remain frozen and only $\pi_\phi$ is updated. Environments sample reference motions from a shared task-family 4D HOI pool, and we apply reference state initialization at every episode reset: the start frame is sampled uniformly from the first 30 frames and clipped to precede the labeled hand-object contact. PPO hyperparameters: actor learning rate $2 \times 10^{-5}$ with an adaptive schedule targeting KL $= 0.01$, critic learning rate $10^{-3}$, discount factor $\gamma = 0.99$, GAE parameter $\lambda_{\mathrm{GAE}} = 0.95$, clipping parameter $\epsilon = 0.2$, entropy coefficient $0.01$, 24 steps per environment, 5 learning epochs with 4 mini-batches, and maximum gradient norm $0.1$. Episodes terminate when the object's $z$-position deviates by more than $0.4$\,m from the reference, the root height deviates by more than $0.25$\,m, or the root orientation error exceeds $1.0$\,rad.

\begin{table}[!t]
    \centering
    \small
    \setlength\tabcolsep{4pt}
    \begin{tabular}{lccc}
        \toprule
        \textbf{Term} & \textbf{Description} & \textbf{Dim} & \textbf{Used} \\
        \midrule
        \multicolumn{4}{l}{\textbf{Proprioception}} \\
        Joint position             & robot joint angles                    & 43 / 29 & Both \\
        Joint velocity             & robot joint angular velocities        & 43 / 29 & Both \\
        Base angular velocity      & in body frame                         & 3       & Both \\
        Previous action            & last meta-action                      & 66 / 29 & Both \\
        Base linear velocity       & in body frame                         & 3       & O    \\
        Gravity direction          & in body frame                         & 3       & S    \\
        \midrule
        \multicolumn{4}{l}{\textbf{Reference motion targets}} \\
        Motion anchor position     & target root position in body frame    & 3       & O    \\
        Motion anchor orientation  & target root orientation 6D            & 6       & O    \\
        Current command            & ref.\ joint position + velocity, current frame & 58 & O \\
        Multi-future command       & ref.\ joint position + velocity, 10 future frames & 580 & O \\
        Multi-future anchor ori    & 10 future root orientations 6D        & 60      & O    \\
        \midrule
        \multicolumn{4}{l}{\textbf{Object state}} \\
        Object position            & current object pos in body frame      & 3       & Both \\
        Object orientation 6D      & current object ori in body frame      & 6       & Both \\
        Object pos delta (10 fut.) & $\mathrm{ref}_{\mathrm{fut}}-\mathrm{sim}_{\mathrm{cur}}$, position           & 30      & Both \\
        Object ori delta (10 fut.) & relative rotation, 6D                 & 60      & Both \\
        Target object position     & reference object pos in base frame    & 3       & O    \\
        Object BPS encoding        & per-object shape descriptor           & 10      & O    \\
        Table position             & in body frame                         & 3       & O    \\
        Table orientation 6D       & in body frame                         & 6       & O    \\
        \midrule
        \multicolumn{4}{l}{\textbf{Hand-object contact}} \\
        Hand-to-object transform   & per right hand (3 pos + 6 ori)        & 9       & O    \\
        Fingertip contact forces   & 3D force per fingertip                & 12      & O    \\
        \midrule
        \multicolumn{4}{l}{\textbf{Scene}} \\
        Local height map           & $11{\times}11$ terrain $z$ in body frame & 121  & S    \\
        \bottomrule
    \end{tabular}
    \vspace{0.5em}
    \caption{\textbf{Policy Observations.} Observations for the object-aware adaptor (O) and the scene-aware tracker (S), grouped by category. When two dims are listed (\eg, 43 / 29), the first applies to the object-aware policy (43-DoF G1 including fingers) and the second to the scene-aware policy (29-DoF G1, no fingers). Proprioceptive and previous-action observations are stored as 10-frame histories; instantaneous values are listed above.}
    \label{tab:observations}
\end{table}

\begin{table}[!t]
    \centering
    \small
    \setlength\tabcolsep{4pt}
    \begin{tabular}{lccc}
        \toprule
        \textbf{Term} & \textbf{Formula} & \textbf{Weight} & \textbf{Used} \\
        \midrule
        \multicolumn{4}{l}{\textbf{Motion tracking reward}} \\
        Anchor position             & $\exp\!\bigl(-\|\widetilde{\bm{p}}^{\mathrm{root}}_t - \bm{p}^{\mathrm{root}}_t\|^2 / \sigma^2\bigr)$ & $0.5$ & Both \\
        Anchor orientation          & $\exp\!\bigl(-\|\widetilde{\bm{r}}^{\mathrm{root}}_t \ominus \bm{r}^{\mathrm{root}}_t\|^2 / \sigma^2\bigr)$ & $2.5$ & Both \\
        Relative body position      & $\exp\!\bigl(-\tfrac{1}{N}\sum_i\|\widetilde{\bm{p}}_{i,t} - \bm{p}_{i,t}\|^2 / \sigma^2\bigr)$ & $1.0$ & Both \\
        Relative body orientation   & $\exp\!\bigl(-\tfrac{1}{N}\sum_i\|\widetilde{\bm{r}}_{i,t} \ominus \bm{r}_{i,t}\|^2 / \sigma^2\bigr)$ & $5.0$ & Both \\
        Body linear velocity        & $\exp\!\bigl(-\tfrac{1}{N}\sum_i\|\widetilde{\bm{v}}_{i,t} - \bm{v}_{i,t}\|^2 / \sigma^2\bigr)$ & $1.0$ & Both \\
        Body angular velocity       & $\exp\!\bigl(-\tfrac{1}{N}\sum_i\|\widetilde{\bm{\omega}}_{i,t} - \bm{\omega}_{i,t}\|^2 / \sigma^2\bigr)$ & $1.0$ & Both \\
        5-point local body          & $\exp\!\bigl(-\tfrac{1}{|\mathcal{S}_5|}\sum_{i\in\mathcal{S}_5}\|\widetilde{\bm{p}}^{\mathrm{loc}}_{i,t} - \bm{p}^{\mathrm{loc}}_{i,t}\|^2 / \sigma^2\bigr)$ & $2.0$ & S \\
        \midrule
        \multicolumn{4}{l}{\textbf{Object reward}} \\
        Object pose tracking        & $w_p\exp\!\bigl(-\alpha_p\|\widetilde{\bm{p}}^{\mathcal{O}}_t - \bm{p}^{\mathcal{O}}_t\|\bigr) + w_r\exp\!\bigl(-\alpha_r\|\widetilde{\bm{r}}^{\mathcal{O}}_t \ominus \bm{r}^{\mathcal{O}}_t\|\bigr)$ & $20.0$ & O \\
        \midrule
        \multicolumn{4}{l}{\textbf{Grasp reward}} \\
        Grasp contact count         & $\min\!\bigl(N^{\mathrm{contact}}_t / N_{\min},\; 1\bigr)$ & $40.0$ & O \\
        Grasp finger direction      & $-\cos(\bm{d}^{\mathrm{thumb}}_t,\, \bm{d}^{\mathrm{index}}_t)\,\mathbb{1}\{C_t\}$ & $10.0$ & O \\
        Grasp contact center        & $\exp\!\bigl(-\gamma\,\tfrac{1}{N_f}\sum_j \|\bm{f}_{j,t} - \bm{c}_t\|\bigr)\,\mathbb{1}\{C_t\}$ & $0.1$  & O \\
        \midrule
        \multicolumn{4}{l}{\textbf{Regularization}} \\
        Latent residual $\ell_2$           & $\|\Delta\bm{z}_t\|^2$                                                                          & $0.1$                 & O    \\
        Finger-primitive limit             & $\mathbb{1}\!\bigl[|a^{\mathrm{fp}}_{j,t}| > 0.5\bigr]$                                         & $-10.0$               & O    \\
        Action rate $\ell_2$               & $\|\bm{a}_t - \bm{a}_{t-1}\|^2$                                                                & $-0.1$                & S    \\
        Anti-shake angular velocity        & $\tfrac{1}{|\mathcal{B}|}\sum_{i\in\mathcal{B}}\bigl[\max(0,\|\bm{\omega}_{i,t}\|-\tau)\bigr]^2$ & $-5{\times}10^{-3}$   & S    \\
        Ankle joint acceleration           & $\sum_{j \in \mathcal{J}_{\mathrm{ankle}}} \ddot{q}_{j,t}^{\,2}$                                & $-2.5{\times}10^{-7}$ & S    \\
        Kinematic reconstruction           & $\|\widetilde{\bm{q}}_t - \mathcal{G}_{\mathrm{rec}}(\bm{z}_t)\|^2$                              & $0.01$                & S    \\
        Joint limit                        & $\sum_j \max\!\bigl(0,\, |q_{j,t}| - q^{\mathrm{lim}}_j\bigr)$                                  & $-10.0$               & S    \\
        Undesired body contact             & $N^{\mathrm{undesired}}_t$                                                                      & $-0.1$                & Both \\
        \bottomrule
    \end{tabular}
    \vspace{0.5em}
    \caption{\textbf{Reward Terms.} Rewards for the object-aware adaptor (O) and the scene-aware tracker (S), organized by the four categories from Sec.~\ref{subsec:physics-tracking}. The rightmost column indicates which tracker uses each term. For grasp terms, the listed weight is the right-hand value; in object-aware training, the left hand uses half the listed weight. $\mathbb{1}\{C_t\}$ is the per-frame motion-label contact indicator that gates the grasp finger-direction and contact-center terms. The object pose-tracking reward is additionally gated by a simulated finger--object contact indicator, so it is active only while the hand is in contact with the object. For anti-shake, $\mathcal{B}=\{$left wrist, right wrist, head$\}$ with deadzone $\tau=1.5$\,rad/s. Per-term Gaussian-kernel bandwidth $\sigma$ varies by term and is given in the released config. Negative weights indicate penalties.}
    \label{tab:rewards}
\end{table}

\subsection{Scene-Aware Tracker}

For scene-level interactions such as stepping over curbs, traversing slopes and stairs, or sitting on chairs, the controller's flat-ground prior is not directly applicable. Instead of attaching a latent adaptor, we fine-tune the controller end-to-end together with a height-map encoder $\epsilon_h$ and an auxiliary kinematic decoder $\mathcal{G}_{\mathrm{rec}}$ on the reconstructed scene-aware data.

\vspace{1mm}\noindent\textbf{Architecture.}
We construct an $11 \times 11$ height map grid centered on the robot with a total extent of $1.5$\,m and a resolution of $0.15$\,m. At each grid point, a downward ray is cast against the scene mesh to obtain the terrain hit position; positions are then transformed into the robot's yaw-aligned local frame, yielding a $(11, 11, 3)$ tensor. The height map is processed by a 3-layer CNN with channels $[64, 128, 256]$, kernel size $3{\times}3$, stride $2$, and LeakyReLU activations; spatial dimensions reduce as $11 \rightarrow 6 \rightarrow 3 \rightarrow 2$, and the output is flattened to a 1{,}024-dim feature vector. This vector is concatenated with the proprioceptive observation and the controller's tokenizer features, then passed through a fusion MLP ($[256]$, SiLU) that produces the latent input to the controller's motion decoder.

\vspace{1mm}\noindent\textbf{Training.}
We train with PPO~\cite{schulman2017proximal} on 64 NVIDIA L40 GPUs with 1{,}024 parallel environments per GPU in Isaac Lab for 30{,}000 iterations. Unlike object-aware tracking, we fine-tune the controller (encoder, FSQ quantizer, and action decoder $\mathcal{G}$) end-to-end together with the height-map encoder $\epsilon_h$ and the parallel kinematic decoder $\mathcal{G}_{\mathrm{rec}}$. $\mathcal{G}_{\mathrm{rec}}$ reconstructs the input motion targets to provide an auxiliary MSE loss (weight $0.01$) that regularizes the latent to remain faithful to the reference. Reference state initialization is sampled uniformly across the full motion at every episode reset. Since hand-object interaction is not involved in these tasks, manipulation-specific reward terms (grasp and object-pose tracking) are disabled. Episodes terminate under cumulative tracking-error thresholds with adaptive strict orientation and foot $xyz$ constraints.

\subsection{Training Cost}
\label{supp:tracking-runtime}
Because each tracking policy is trained jointly over a shared task-family pool rather than fit per sequence, we report the amortized training cost per motion, defined as the total training wall-clock of a run divided by the number of motions in its pool. A full policy is trained for $30{,}000$ PPO iterations on $64$ NVIDIA L40 GPUs with $1{,}024$ environments per GPU, which takes roughly $30$ hours, and each run trains $2{,}000$--$4{,}000$ motions jointly. The amortized cost is therefore only about $0.5$--$0.9$ minutes per motion, far below the per-sequence cost of fitting a controller to each trajectory in isolation. As the pool grows with additional in-family motions, we do not retrain from scratch: we warm-start from the current policy and fine-tune, which typically converges within $6{,}000$ iterations (about one fifth of a full run, $\sim$6 hours), reducing the amortized cost of incorporating new motions to roughly one fifth of the full-training figure.

\section{Experiment Details}

\subsection{Human-Object Interaction Generation}
\label{supp:additional-results}

\noindent\textbf{Baselines.}
We compare against training-based and training-free 4D HOI generation approaches. Training-based baselines include CHOIS~\cite{li2024controllable}, a controllable motion generation framework guided by language and sparse waypoints, and HOIDiff~\cite{peng2025hoi}, a diffusion-based model for affordance-conditioned HOI synthesis. For training-free comparison, we evaluate DAViD~\cite{kim2025david}, which generates the first frame using an image generative model~\cite{labs2025flux1kontextflowmatching} and produces the video from that frame. Since image generation often fails under partial control signals (\eg, Canny edge maps), we generate 24 images per object and manually select a successful result as the starting frame. To ensure fairness, DAViD uses Kling's image-to-video model under the same setting as our approach.

\vspace{1mm}\noindent\textbf{User Study.}
Beyond the quantitative metrics reported in Sec.~\ref{subsec:hoi-generation-results}, we conduct a user study with 30 participants to assess perceptual quality. In each trial, participants view sequences from three of the four methods drawn at random and select the result with the most appropriate object affordances (Aff.\ Real.) and the most physically plausible motion (Phys.\ Real.). As shown in Table~\ref{tab:user-study}, \METHOD is preferred by a wide margin on both criteria. This perceptual study complements, but does not replace, the physics-based tracking and robot-execution metrics reported in Sec.~\ref{subsec:hoi-generation-results} and Sec.~\ref{subsec:deployment-results}.
\begin{table}[!t]
    \centering
    \small
    \setlength\tabcolsep{8pt}
    \begin{tabular}{lcc}
        \toprule
        \textbf{Methods} & Aff. Real. $\uparrow$ & Phys. Real. $\uparrow$ \\ \midrule
        HOIDiff~\cite{peng2025hoi}      & 2.0\%           & 1.9\%           \\
        CHOIS~\cite{li2024controllable} & 12.2\%          & 16.8\%          \\
        DAViD~\cite{kim2025david}       & 11.2\%          & 10.4\%          \\
        \midrule
        Ours                            & \textbf{74.7\%} & \textbf{70.9\%} \\
        \bottomrule
    \end{tabular}
    \caption{\textbf{User Study.} 30-participant pick rates for the most appropriate object affordances (Aff.\ Real.) and the most physically plausible motion (Phys.\ Real.) on the shared 20-object evaluation set. Pick rates have a theoretical upper bound of 75\% under 3-of-4 random sampling.}
    \label{tab:user-study}
\end{table}

\begin{figure*}[!t]
    \centering
    \includegraphics[width=0.99\textwidth]{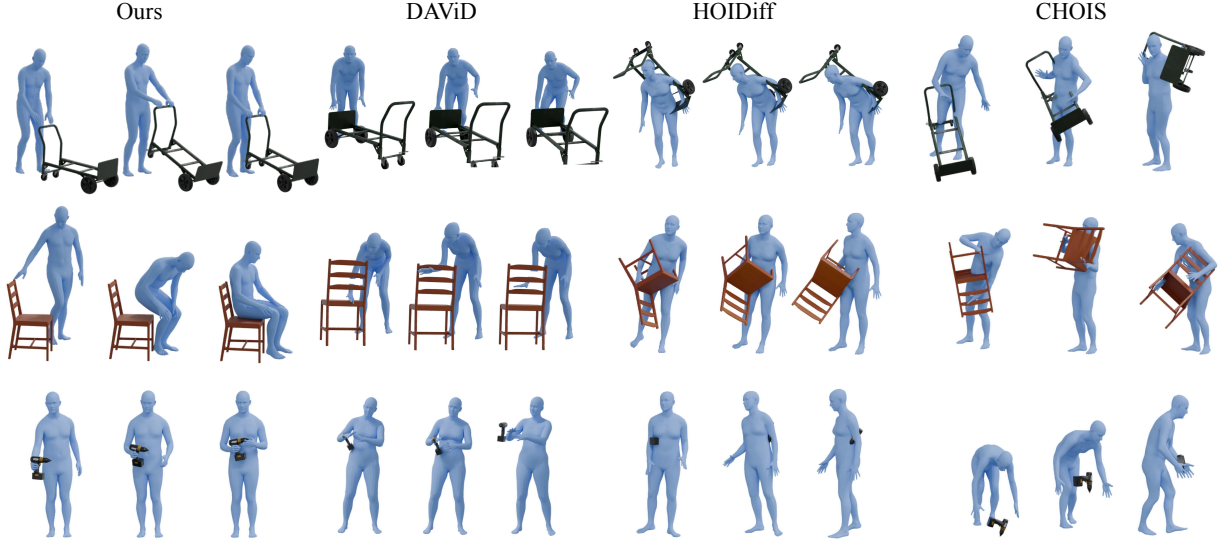}
    \caption{\textbf{Qualitative Comparison.} Representative 4D HOI sequences from each method on the shared 20-object evaluation set. \METHOD produces more coherent motions with accurate contact and natural hand poses, while baseline methods often yield unrealistically flat or static hand configurations.}
    \label{fig:qualitative_comparison}
\end{figure*}

\vspace{1mm}\noindent\textbf{Qualitative Comparison.}
Fig.~\ref{fig:qualitative_comparison} shows that \METHOD produces more coherent motions with accurate contact and natural hand poses, while baseline methods often yield unrealistically flat or static hand configurations unsuitable for downstream humanoid skill learning.

\begin{table*}[!t]
    \centering
    \small
    \setlength\tabcolsep{4pt}
    \begin{tabular}{lccccccccc}
        \toprule
         & \multicolumn{5}{c}{\textbf{Reconstruction Quality}} & \multicolumn{3}{c}{\textbf{Tracking Quality}} \\
        \cmidrule(lr){2-6} \cmidrule(lr){7-9}
        \textbf{Methods}
        & Contact $\downarrow$ & Pen. $\downarrow$ & MPJPE $\downarrow$ & Human Smo. $\downarrow$ & Obj Smo. $\downarrow$
        & SR $\uparrow$ & ObjPos $\downarrow$ & MPJPE-L $\downarrow$ \\ \midrule
        w/o $L_{\text{proj}}$  & 0.016 & 1.93\% & 16.70 & 0.0023 & 0.0011 & 41.6\%          & 0.374          & 47.1          \\
        w/o $L_{\text{depth}}$ & 0.017 & 1.97\% &  4.35 & 0.0020 & 0.0011 & 42.6\%          & 0.372          & 49.3          \\
        w/o $L_{\text{cont}}$  & 0.024 & 1.52\% &  4.81 & 0.0022 & 0.0009 & 53.3\%          & 0.332          & 52.4          \\
        \midrule
        Ours (Full)            & 0.015 & 1.81\% &  4.89 & 0.0020 & 0.0009 & 81.4\%          & 0.135          & 41.8          \\
        \bottomrule
    \end{tabular}
    \caption{\textbf{Ablation Study on Reconstruction Losses.} Reconstruction quality (Contact, Pen., MPJPE in pixel space, motion smoothness) and downstream tracking quality (SR, ObjPos, MPJPE-L) for each loss ablation. Each loss targets a different failure mode in reconstruction; the full model achieves the best downstream tracking despite not minimizing every reconstruction proxy in isolation.}
    \label{tab:ablation_recon_suppl}
\end{table*}

\noindent\textbf{Reconstruction Ablation.}
We ablate the interaction-aware reconstruction losses ($L_{\text{proj}}$, $L_{\text{depth}}$, $L_{\text{cont}}$) on the 124-motion benchmark. As shown in Table~\ref{tab:ablation_recon_suppl}, removing $L_{\text{proj}}$ degrades image-space consistency, while removing $L_{\text{depth}}$ or $L_{\text{cont}}$ weakens metric interaction quality. These reconstruction-level errors propagate to downstream tracking: each ablation substantially reduces tracking success rate and increases trajectory deviation, while the full model achieves the best overall downstream tracking quality.

\end{document}